\documentclass[runningheads]{llncs}
\usepackage{graphicx}
\usepackage{tikz}
\usepackage{comment}
\usepackage{amsmath,amssymb} % define this before the line numbering.
\usepackage{color}
\usepackage{hyperref}
\usepackage{booktabs}
\usepackage{footmisc}

\usepackage[accsupp]{axessibility}  % Improves PDF readability for those with disabilities.

\begin{document}
% \linenumbers
\pagestyle{headings}
\mainmatter
\def\ECCVSubNumber{1952}  % Insert your submission number here

\title{Learning Invariant Visual Representations for Compositional Zero-Shot Learning}

\begin{comment}
% \titlerunning{ECCV-22 submission ID \ECCVSubNumber} 
% \authorrunning{ECCV-22 submission ID \ECCVSubNumber} 
% \author{Anonymous ECCV submission}
% \titlerunning{Learning Invariant Visual Representations for Compositional Zero-Shot Learning} 
\end{comment}
%******************

% \begin{comment}
\titlerunning{Learning Invariant Visual Representations for CZSL}
\author{Tian Zhang\inst{1\thanks{Equal contribution; codes are available at \url{https://github.com/PRIS-CV/IVR}.}} \and
Kongming Liang\inst{1\footnotemark[1]} \and
Ruoyi Du\inst{1} \and
Xian Sun\inst{2} \and
Zhanyu Ma\inst{1} \and
Jun Guo\inst{1}} 

% Equal contribution
%
\authorrunning{Zhang et al.}
\institute{Pattern Recognition and Intelligent System Laboratory, School of Artificial Intelligence, Beijing University of Posts and Telecommunications \\
\email{\{zhangtian1874,liangkongming,duruoyi,mazhanyu,guojun\}@bupt.edu.cn}
\and Aerospace Information Research Institute, Chinese Academy of Sciences \email{sunxian@aircas.ac.cn}
}

% \end{comment}
%******************

\maketitle

\begin{abstract}
Compositional Zero-Shot Learning (CZSL) aims to recognize novel compositions using knowledge learned from seen attribute-object compositions in the training set.
Previous works mainly project an image and a composition into a common embedding space to measure their compatibility score. 
However, both attributes and objects share the visual representations learned above, leading the model to exploit spurious correlations and bias towards seen pairs.
Instead, we reconsider CZSL as an out-of-distribution generalization problem.
If an object is treated as a domain, we can learn object-invariant features to recognize the attributes attached to any object reliably. Similarly, attribute-invariant features can also be learned when recognizing the objects with attributes as domains. 
Specifically, we propose an invariant feature learning framework to align different domains at the representation and gradient levels to capture the intrinsic characteristics associated with the tasks.
Experiments on three CZSL benchmarks demonstrate that the proposed method significantly outperforms the previous state-of-the-art.

\end{abstract}

\section{Introduction}
Humans can easily generalize the \emph{red} state from \emph{apples} to \emph{tomatoes} even if no images of \emph{red tomatoes} have been seen. Since visual concepts follow the long tailed distribution, the instances of most concepts are rarely presented in the real world scenario. %加引用
%it was possible to provide more data for the tail classes and just ignore the open classes that might appear in the testing dataset
Therefore, the ability to generalize the learned knowledge to novel concepts is of vital importance for human to recognize a large number of concepts and is considered as one of the hallmarks of human intelligence~\cite{2017From,2019Task}. The goal of Compositional Zero-Shot Learning (CZSL) is to build a model that can learn the attributes and objects from seen compositions and generalize them well to unseen compositions.
For instance, the model trained with images of \emph{red apples} and \emph{green tomatoes} can correctly predict images of \emph{red tomatoes}.

\begin{figure}[t]
\centering
\includegraphics[height=5.2cm]{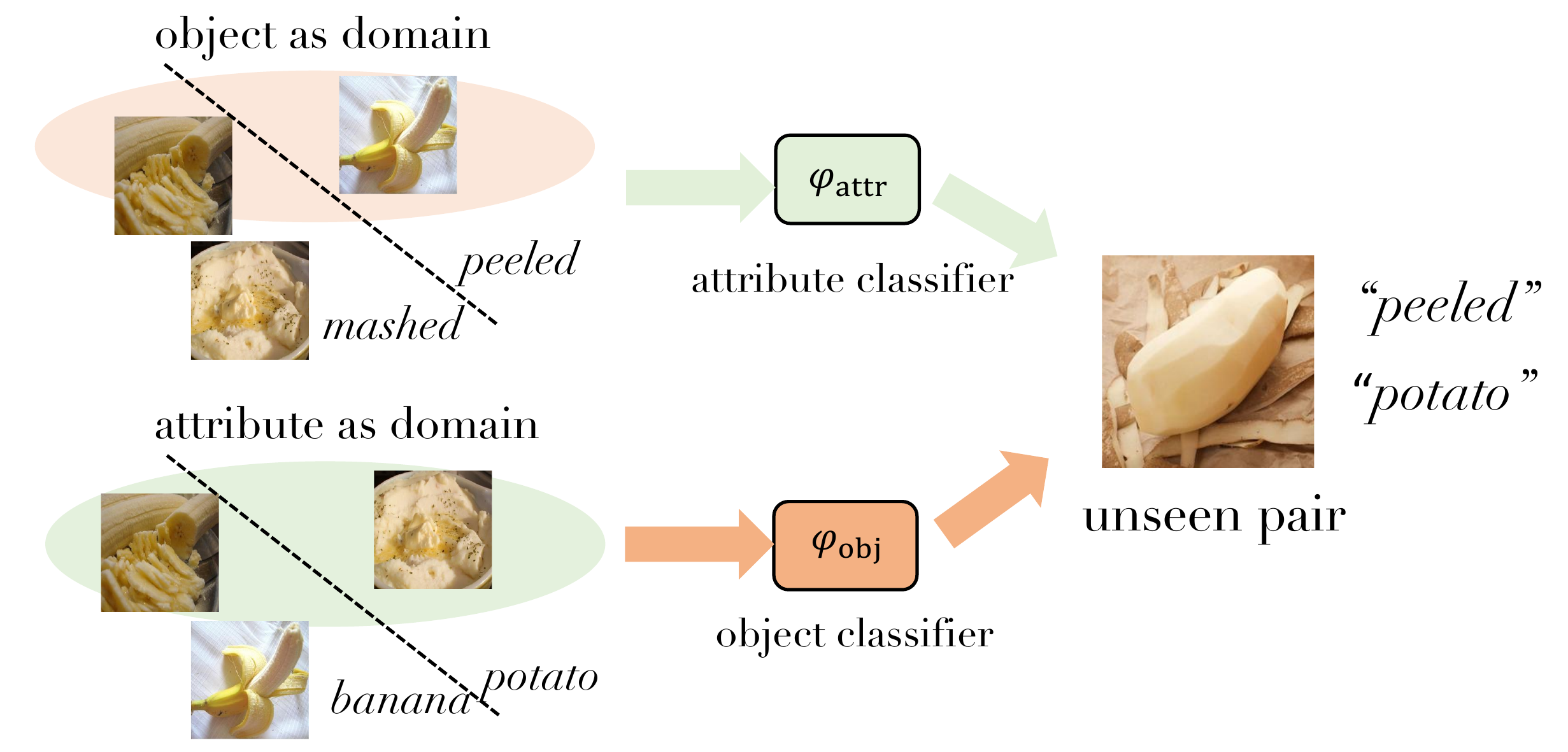}  % 5 5.2
\caption{The illustration of our motivation. Ellipses represent corresponding domains.
The samples in one ellipse belong to the same domain, the samples outside the ellipse belong to other domains.
And the dotted lines represent category decision boundaries within decoupled feature space.
}
\label{fig:schematic}
\end{figure}

%Given the diversity of attributes and objects, it is difficult to learn all possible compositions of the visual concepts.
Previous works \cite{2017From,2018Attributes,2020Symmetry,2021Open} in CZSL mainly project image features and attribute-object composition features into a common embedding space and constrain the features belonging to the same concept to be closer. 
Specifically, the current state-of-the-art method~\cite{2021Open} use cosine similarity to calculate the compatibility score of images and compositions in the embedding space.
Since the features are learned in a composition way, they are not disentangled for attribute and object which makes the model over-rely on a limited number of attribute-object pairs in the training process.
For instance, when machines had only seen \emph{red apples}, they might easily misidentify \emph{red tomatoes} as \emph{red apples} since classifier had prone to spuriously correlate \emph{red} with \emph{apple}.
Machine learning models are data-driven and typically require samples of various perspectives and lighting. This makes them often rely on spurious features~\cite{arjovsky2019invariant,shahtalebi2021sand,zhang2021towards,khezeli2021invariance,ahuja2021invariance} unrelated to the core concept and lose generalization performance~\cite{geirhos2018imagenet}, especially in zero-shot learning scenarios.
% On the other hand, learning attributes and objects independently may lose some of their relevance, compromising the accuracy of seen compositions. Therefore, leveraging a balance between “dependence” – recognize attributes and objects in an entangled way, and “independence” – recognize attributes and objects in a decoupling way may actually assist the model in achieving better performance~\cite{ruis2021independent}.
Therefore, recognizing attributes and objects independently may actually assist the model in achieving better performance.

In this paper, we leverage the idea of Domain Generalization (DG) to improve the ability of the model to generalize to unknown compositions. 
Most deep learning methods work well under the \emph{i.i.d.} assumption that training and testing data are independently and identically distributed~\cite{peng2019domain,bengio2019meta}. However, this assumption does not always hold true in reality. When the probability distributions of training and testing data are different, the performance of deep learning models is often degraded due to the domain shift~\cite{quinonero2008dataset,2021Generalizing}.
DG trains model only with data from the source domain, making it generalize well to the unseen arbitrary target domain. For instance, given a training set consisting of \emph{photos}, \emph{cartoon images} and \emph{paintings}, DG requires training a model to have promising performance in classifying \emph{sketches}, which are significantly different from the images in the training set. Most of the work alleviates domain shift by aligning feature distributions of the source with target domains, resulting in domain-invariant features.

Since a domain is composed of data that are sampled from a distribution~\cite{2021Generalizing}, the Compositional Zero-Shot Learning task is analogous to two DG sub-tasks in essence, by taking objects or attributes as domains. 
As shown in the Figure~\ref{fig:schematic}, in the case of treating objects as domains, if the model learns the attributes of \emph{mashed} and \emph{peeled} in the \emph{banana} domain, then we expect that it can also reliably recognize the attribute of \emph{peeled} when generalized to the \emph{potato} domain. Similarly, in the case of treating attributes as domains, if the model learns the objects of \emph{banana} and \emph{potato} in the \emph{mashed} domain, it should recognize the object of \emph{potato} when generalized to the \emph{peeled} domain. 
Eventually, the model is able to successfully recognize the unseen pairs (\emph{peeled potato}).
We simulate a domain generalization scenario by designing a triplet input network. 
To decouple the highly-coupled features, we construct two branches, the object-domain branch and the attribute-domain branch.
For the object-domain branch, our goal is to accurately recognize the attribute regardless of object labels.
We learn consistency at the representation level by discarding object-specific channels. Moreover, we minimize the gradient differences of attribute prediction in different object domains to achieve gradient-level consistency.
For the attribute-domain branch, we learn attribute-invariant features in the same way.
%We explicitly promote agreement of gradients within different object domains to ensure that cross-domain invariance is captured.
%And for the attribute-domain branch, we learn the attribute-domain invariant feature of the object based on the alignment of gradients between different attribute domains. 
Finally, by penalizing domain-specific power of features, we discover invariant mechanisms in the data which are hard to vary across examples and thus learn the optimal attribute classifier and object classifier.

The contributions of the paper are summarized below. 
(1) To the best of our knowledge, we are the first to solve the Compositional Zero-Shot Learning task from a Domain Generalization perspective. In other words, the compositional learning problem is transformed into a domain-shift problem. 
(2) We treat attributes or objects as domains and align different domains to learn domain-invariant features, thus improving the generalization performance of the model to recognize unseen pairs.
(3) We prove the effectiveness of our method through abundant experiments.

\section{Related works}
\subsection{Compositional Zero-Shot Learning}
Compositional Zero-Shot Learning (CZSL) is a special case of Zero-Shot Learning (ZSL)~\cite{palatucci2009zero,xian2017zero,xian2018feature}. Given a training set containing a set of attribute-object compositions, CZSL aims to recognize unknown compositions of these attributes and objects at inference time.
Part of the work proposes to learn classifiers for individual concepts and combine them to recognize integrated concepts.
Chen et al.~\cite{chen2014inferring} deduce unobserved attribute-object pairs through tensor decomposition during training.
Misra et al.~\cite{2017From} consider compositionality and contextuality as the key to solving CZSL, and they merge classifiers for primitive concepts into classifiers for composite concepts.
A most popular line of work involves embedding attribute-object compositions into a feature space.
Nagarajan et al.~\cite{2018Attributes} argue that objects are entities while attributes are properties of the objects and consider the composition of attributes and objects as a learned transformation.
%Nan et al.~\cite{nan2019recognizing} propose a generative model with the encoder-decoder mechanism to learn an intrinsic attribute-object representation. 
Wei et al.~\cite{wei2019adversarial} model the attribute-object relationships within the feature space based on a GAN framework.
Li et al.~\cite{2020Symmetry} propose symmetry as an essential principle for attribute-object transformations and introduce group theory as an axiomatic foundation to satisfy the specific principles of nature. 
Mancini et al.~\cite{2021Open} propose a new open world setting for CZSL task where the prior knowledge of unseen compositions is not provided. 
%They also introduce a feasibility assessment mechanism to score compositions in order to detect attributes and objects from images. 
%All the work mentioned above is to project image features and text features into a common embedding space and constrain the features belonging to the same compositions to be closer. If attributes and objects are not detected in a decoupled way, the model will over-rely on a limited number of attribute-object pairs in the training process, so the model will be difficult to generalize to unseen compositions.
Instead, other works learn the joint compatibility between the input image and the attribute-object pair.
Purushwalkam et al.~\cite{2019Task} train a set of network modules jointly with a gating network to produce features that indicate compatibility between the input image and the concept.
Atzmon et al.~\cite{atzmon2020causal} describe CZSL from a causal perspective and try to find which intervention cause the image. 
%They build an embedding-based architecture that looks for causally stable representations for composition recognition.
%A lot of previous works emphasize that the expression of attributes depends on objects to a certain extent.
Unlike these works, we focus on the independence between the sub-concepts and learn an attribute classifier and object classifier that can be generalized to new compositions.

\subsection{Domain Generalization}
In reality, the distribution of training and test sets is often different, leading to model performance degradation. This problem is known as out-of-distribution generalization or domain generalization~\cite{blanchard2011generalizing,muandet2013domain,2021Domain}.
%Domain generalization (DG) aims to learn a model that generalizes knowledge learned from the source domains to the target domains, which are generally out of the training distribution~\cite{blanchard2011generalizing,muandet2013domain,2021Domain}.
Since the generalization ability of the model often depends on the quantity and quality of training data~\cite{2021Generalizing}, one line of work increases the diversity of existing training data through data augmentation and data generation to learn more general representations. 
Qiao et al.~\cite{Qiao_2020_CVPR} leverage Wasserstein Auto-Encoders (WAE)~\cite{tolstikhin2017wasserstein} to help generate samples that retain semantics and have large domain transportation.
Shankar et al.~\cite{Shankar2018GeneralizingAD} introduce a domain classifier to expand the training data by disturbing the input data.
Carlucci et al.~\cite{2020Jigsaw} enrich the understanding of the data by solving puzzle problems, allowing the model to induce invariance and regularity autonomously. 
A different line of work uses domain alignment techniques or feature disentanglement to learn domain-invariant features.
Sun et al.~\cite{sun2016deep} conduct domain alignment by matching the mean and variance of representations in different domains.
Li et al.~\cite{2018DomainAdver} use Maximum Mean Discrepancy (MMD) to align different domains to obtain domain-invariant representation.
Peng et al.~\cite{peng2019domain} decouple features into domain-invariant features, domain-specific features, and class-irrelevant features through adversarial learning. 
Huang et al.~\cite{2020RSC} propose a self-challenge mechanism, which iteratively discards the dominant features activated on the training data.
%and forces the network learning domain-invariant representation.
Kim et al.~\cite{kim2021selfreg} propose self-supervised contrastive regularization to map the latent representations of the positive pair samples close together.
In this paper, we mainly leverage the idea of exploring invariance in DG to enhance the performance of the CZSL task.

\section{Methods}
\begin{figure}[t]
\centering
\includegraphics[height=5cm]{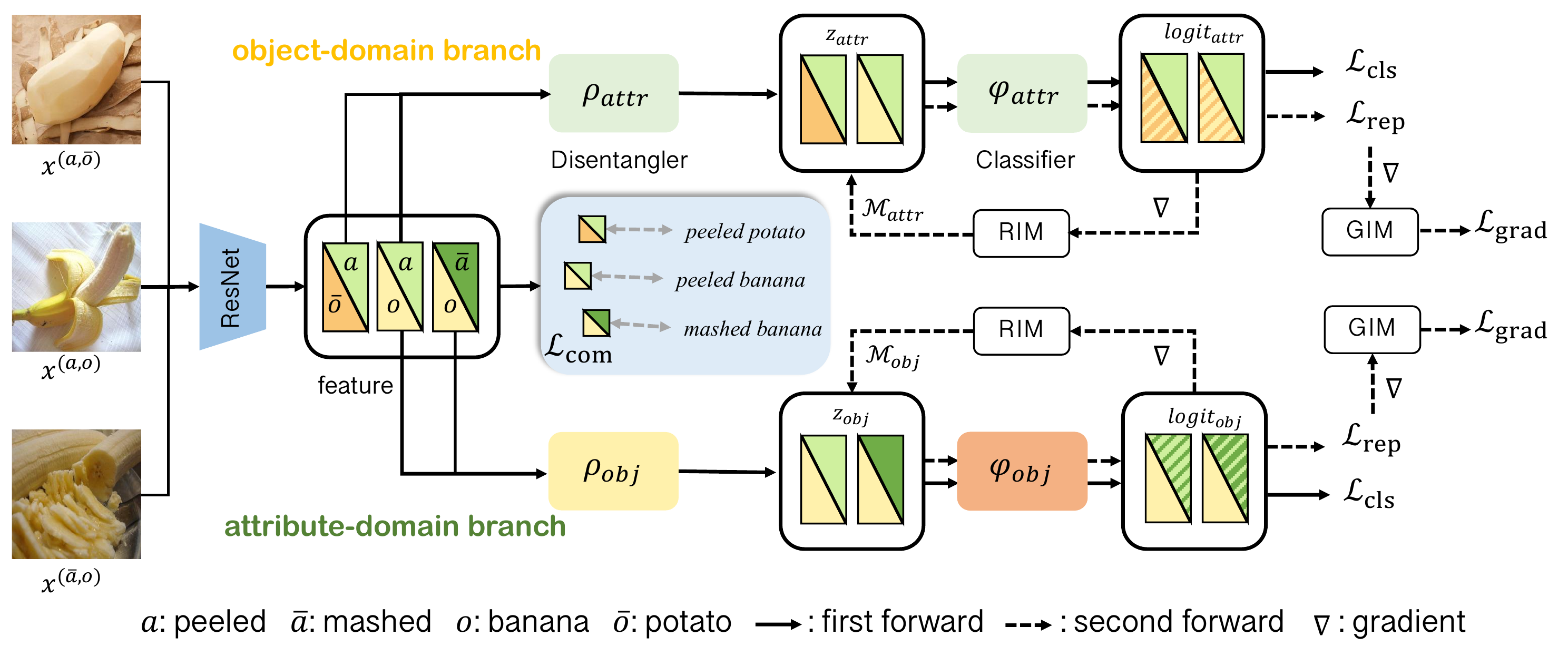}
\caption{Overview of the proposed framework. We construct object-domain branch and attribute-domain branch. In the object-domain branch, we execute consistent alignment across different object domains so that the model learns the essential characteristics of the attribute. The same for the attribute-domain branch.
}
\label{fig:method}
\end{figure}

\begin{figure}[t]
\centering
\includegraphics[height=2.3cm]{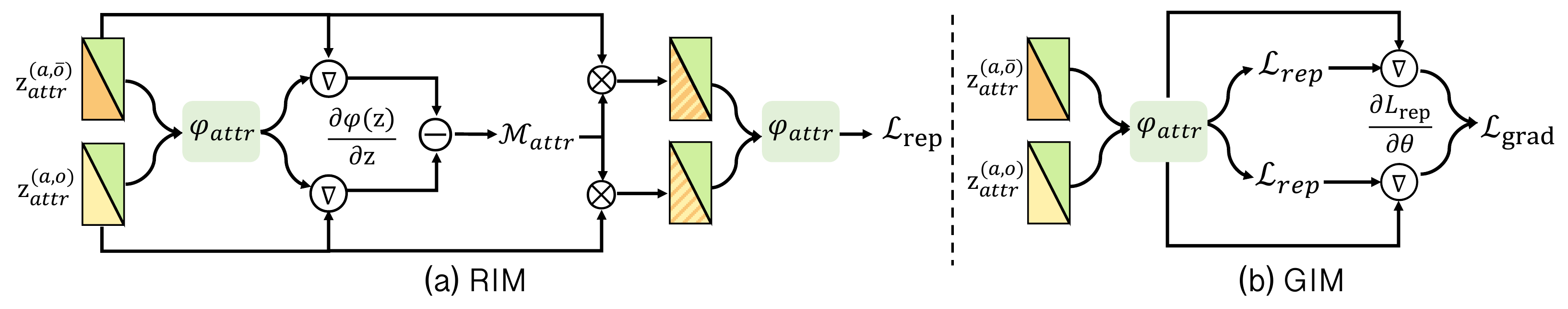}
\caption{Consider the object-domain branch with the representations $z^{(a,o)}_{attr}$ and $x^{(a,\bar{o})}_{attr}$. 
(a) In the representation invariant mechanism (RIM), we learn object-irrelevant attribute features at the representation level by filtering out object-specific channels.
(b) In the gradient invariant mechanism (GIM), we learn object-irrelevant attribute features at the gradient level by minimizing the distance between gradients of different object domains.}
\label{fig:sub-method}
\end{figure}

 In typical CZSL, we have access to all the attributes and objects, while only part of attribute-object compositions can be obtained in the training phase. The goal is to recognize unknown compositions of individual attribute and object concepts. Composing the learned knowledge into unseen compositions heavily relies on out-of-distribution generalization ability~\cite{koyama2020out,krueger2021out,shahtalebi2021sand,khezeli2021invariance,ahuja2021invariance}.
Therefore, we formalize the CZSL problem into two domain generalization sub-tasks, in which we consider attributes as domains to recognize objects and vice versa.
Then, two types of invariant mechanisms are proposed to remove the spurious domain-specific features and improve the generalization ability of the model.
%learn domain-invariant features

%To mitigate the difficulty of alignment, we penalize domain-specific prediction capabilities by discarding highly differentiated channels based on gradients.
An overview of our proposed framework is shown in Figure~\ref{fig:method}. In the following sections, we first introduce the visual and composition embedding learning procedure. Then we present how the visual features are decomposed and processed by representation and gradient invariant mechanisms in sequential. Finally, we describe the training and inference methodologies.

\subsection{Visual and Composition Embedding}
We need to train a model that learns a mapping from a set of images $X$ to a set of compositions $Y=Y_{attr}\times Y_{obj}$, 
%$Y_i$ ($i\in\left\{attr, obj \right\}$), 
where $Y_{attr}$ is a set of attribute labels and $Y_{obj}$ is a set of object labels. 
The composition label is divided into $Y=Y_s\cup Y_u$, where $Y_s$ is the set of seen compositions during training and $Y_u$ is the set of unseen compositions for the validation and test sets, with $Y_s\cap Y_u=\emptyset$.
Given an image $x\in X$ in the training set and  its corresponding label $y\in Y_s$, we first use a pre-trained network $f(\cdot )$ (e.g., ResNet-18~\cite{he2016deep}) to extract its visual embedding. Then, the composition embedding function $g(\cdot)$ projects the combined concepts $y$ into a common semantic space.
The composition classification loss can be obtained by minimizing the distance between the two embedding features,
\begin{equation}
h_{comp}(x,y)=d_{cos}(f(x), g(y)),
\end{equation}
where $d_{cos}(\cdot, \cdot)$ is the cosine distance of the input two embeddings. The distance in the embedding space represents the compatibility of the input image and the attribute-object composition. 
Therefore, the smaller the distance is, the higher probability that the composition exists in the image~\cite{2018Attributes}.

However, the visual representations learned in the above manner are shared by both attributes and objects, which may lead the model exploiting spurious correlations and bias the model against seen pairs. %and compromise the generalization performance.
In this work, we utilize invariant feature learning to decouple attributes and objects from a domain generalization perspective. The learned invariant features explore the independence between attribute and object concepts and prove to be effective to complement the conventional visual embedding. 

\subsection{Decomposing Visual Features}
In order to conduct the invariant feature learning for CZSL, we need to decompose the visual features into two parts by considering object and attribute as domain respectively. 
%We denote $a$, $o$, $\bar{a}$, and, $\bar{o}$ as labels of the attributes and objects and 
Here, we design a triplet input network with $x^{a,\bar{o}}$, $x^{a,o}$ and $x^{\bar{a},o}$ as inputs to diversify the inter-domain variation, 
where $a, \bar{a} \in Y_{attr}$ denote different attributes, and $o, \bar{o} \in Y_{obj}$ denote different objects,
%$(y_a,y_o)$, $(y_a,y_{\bar{o}})$ and $(y_{\bar{a}},y_o^{(a,o)})$ 
%$(y_a^{(a,o)},y_o^{(a,o)})$, $(y_a^{(a,\bar{o})},y_{o}^{(a,\bar{o})})$ and $(y_{a}^{(\bar{a},o)},y_o^{(\bar{a},o)})$ respectively. 
%The $a$, $o$, $\bar{a}$, and, $\bar{o}$ denote labels of the attributes and objects.
%And the $\mathfrak{a}$ and $\mathfrak{o}$ represent attribute and object semantic concepts to be recognized.
%The symbol $\ \bar{}\ $ denotes a negative label, 
e.g., $x^{(a,\bar{o})}$ represents an image of different object with the same attribute as $x^{(a,o)}$. We denote the composition set of a triplet input as $\mathcal{C}=\left\{ (a,\bar{o}),(a,o),(\bar{a},o)\right\}$. And the classification task set is denoted as $\mathcal{T} = \left\{attr, obj \right\}$.

The extracted visual features from the pre-trained network $f(\cdot )$ are directly fed into two individual MLPs, attribute disentangler $\rho_{attr}(\cdot )$ and object disentangler $\rho_{obj}(\cdot )$. For $i\in \mathcal{C}$ and $j\in \mathcal{T}$, the image features of $x^i$ can be decoupled as $z_j^i=\rho_j(f(x^i))$.
Given the cross entropy loss function $l(\cdot, \cdot)$, the attribute and object classification loss can be defined as, 
\begin{equation}
L_{cls}=\sum_{i\in \mathcal{C}}\sum_{j \in \mathcal{T}}l(\varphi_j(z_j^i;\theta_{j}),y_j^i ),
\end{equation}
where $\varphi _{j}(\cdot )$ denotes the classifier of task $j$ which predict classification labels over the decomposed visual features. $\theta_{j}$ represents the parameters of classifier $\varphi_{j}(\cdot)$.

\subsection{Learning Invariant Features for CZSL}
% 加过渡
A notion of invariance implies something that stays the same while something else changes~\cite{parascandolo2020learning}. 
Capturing invariance helps model learn the core features related to the label. 
Returning to the previous example, the explanations to distinguish \emph{tomatoes} from \emph{apples} should be invariant, no matter whether the \emph{tomatoes} are \emph{red} or \emph{green}.
Therefore, we leverage invariant feature learning to capture the invariance of objects when attributes change or vice versa. 
Finally the learned invariant features of attributes and objects can be generalized to novel compositions.

When the model takes $x^{(a,o)}$ and $x^{(a,\bar{o})}$ as inputs, we construct a scenario that recognize attribute concept with object as domain. Similarly, with $x^{(a,o)}$ and $x^{(\bar{a},o)}$ as inputs, we construct a scenario that recognize object concept in terms of attribute as domain.
Our goal is to recognize an attribute associated with any objects and recognize an object described by any attributes.
To improve the generalization performance of the model, we explicitly promote invariance to disentangle spurious features at representation level and gradient level.

%Consider the case where the train dataset consists of $S = 2$ object domains $\mathcal{D}_{obj}=\left \{ \mathcal{D}_{o}, \mathcal{D}_{\bar{o}} \right \} $ for the inputs $x^{(a,o)}$ and $x^{(a,\bar{o})}$. 
%Our goal is to mitigate the distribution shift between object domains and capture the object-irrelevant features of attribute in order to accurately recognize the attribute $a$ over the two object domains.
%We explicitly promote invariance to disentangle spurious features at representation level and gradient level.

%\vspace{1.5mm} 
\noindent{\bf Representation Invariant Learning.}
To learn an invariant classifier that helps with generalizing to new domain, we explore invariance at the representation level to pull together samples with the same class from different domains in the feature space. In other words, learn a model that maps different domains to a single statistical distribution~\cite{arjovsky2019invariant,shahtalebi2021sand}.

Firstly, we calculate the gradient of prediction results over the different domains with respect to the representation,
%\begin{equation}g_{\mathcal{D}_{o}}=\frac{\partial(\varphi_a(z_a^{a,o};\theta _a)\odot y_a)}{\partial z_a^{a,o} },\end{equation}
%\begin{equation}g_{\mathcal{D}_{\bar{o}}}=\frac{\partial(\varphi_a(z_a^{a,\bar{o}};\theta _a)\odot y_a)}{\partial z_a^{a,\bar{o}} },\end{equation}
\begin{equation}g_j^i=\frac{\partial([\varphi_j(z_j^{i};\theta _j)]^\top \cdot y_j^i)}{\partial z_j^i}.\end{equation}
%where $\odot$ denotes an element-wise product.

The representations associated with the similar gradients indicate intrinsic characteristic of attribute concepts that are invariant to object factors or vice versa.
Thus we calculate the absolute value of the difference between the two gradients,
%\begin{equation}\Delta g_{\mathcal{D}_{obj}}=   \left| g_{\mathcal{D}_{o}}-g_{\mathcal{D}_{\bar{o}}} \right|.\end{equation}
\begin{equation}
\Delta g_{attr}= \left| g_{attr}^{(a,o)}-g_{attr}^{(a,\bar{o})} \right|,\  \Delta g_{obj}=   \left| g_{obj}^{(a,o)}-g_{obj}^{(\bar{a},o)} \right|.
\end{equation}

The semantic channels with small difference can be regarded as object-invariant feature channels of attribute and attribute-invariant feature channels of object. 
We sort the difference from the largest to the smallest, taking the value at $\alpha$ percent, and denoted as $t^{\alpha}$. Then we construct a mask that shares the same dimension with the representation as follows. For the $k^{th}$ element,
%\begin{equation}m_{\mathcal{D}_{obj}} =\begin{cases}
%  0, & \text{ if } \Delta g_{\mathcal{D}_{obj}}(k)\ge = t^{\alpha}\\
%  1, & \text{ else } \label{con:mask}\end{cases}.\end{equation}
\begin{equation}m_{j}(k) =\begin{cases}
  0, & \text{ if } \Delta g_{j}(k)\ge t^{\alpha}\\
  1, & \text{ else } \label{con:mask}\end{cases}.
  \end{equation}

By overwriting the mask to the original representation, the network filters out domain-specific feature channels to learn the domain-invariant feature,
%\begin{equation}\hat{z}_a^{a,o} =z_a^{a,o}\odot m_{\mathcal{D}_{obj}},\quad \hat{z}_a^{a,\bar{o}} =z_a^{a,\bar{o}}\odot m_{\mathcal{D}_{obj}}.\end{equation}
\begin{equation}
\hat{z}_j^{i} =z_j^{i}\odot m_{j}.
\end{equation}

Then we computes the cross entropy loss with the object-irrelevant attribute-specific representation and the attribute-irrelevant object-specific representation,
%\begin{equation}L_{rep^a}^i=l(\varphi_a(\hat{z}_a^{a,o};\theta_a ),y_a),\quadL_{rep^a}^i=l(\varphi_a(\hat{z}_a^{a,\bar{o}};\theta_a ),y_a).\end{equation}
%\begin{equation}
%L_{rep}=\sum_{i}\sum_{i\in j}l(\varphi_i(\hat{z}_i^{j};\theta_i ),y_i^j).
%\end{equation}

\begin{equation}
\begin{aligned}
L_{rep}&=l(\varphi_{attr}(\hat{z}_{attr}^{(a,o)};\theta_{attr} ),y_{attr}^{(a,o)})+l(\varphi_{attr}(\hat{z}_{attr}^{(a,\bar{o} )};\theta_{attr} ),y_{attr}^{(a,\bar{o} )}) \\
&+l(\varphi_{obj}(\hat{z}_{obj}^{(a,o)};\theta_{obj} ),y_{obj}^{(a,o)})+l(\varphi_{obj}(\hat{z}_{obj}^{(\bar{a},o )};\theta_{obj} ),y_{obj}^{(\bar{a},o )}).
\end{aligned}
\end{equation}

% \begin{equation}
% \begin{aligned}
% L_{rep}&=\sum_{p\in Y_{obj}}l(\varphi_{attr}(\hat{z}_{attr}^{(a,p)};\theta_{attr} ),y_{attr}^{(a,p)})+
% \sum_{q\in Y_{attr}}l(\varphi_{obj}(\hat{z}_{obj}^{(q,o)};\theta_{obj} ),y_{obj}^{(q,o)}).
% \end{aligned}
% \end{equation}

%\vspace{1.5mm} 
\noindent{\bf Gradient Invariant Learning.}
Since reducing empirical risk~\cite{zhang2017mixup} across different domains can reduce the sensitivity of models to distribution shift~\cite{krueger2021out}, we execute gradient-level domain alignment to optimize different domains in the same direction, which will penalize the network to minimize the dispersion of gradients in different domains to capture invariance. 
The objective of enhancing gradient consistency is to find local or global minimum in the loss space across all of the training domains and let the network share similar Hessians for different domains~\cite{shahtalebi2021sand}.
% 加过渡

We calculate the gradient of attribute prediction results to attribute classifier in different object domains as well as the gradient of object prediction results to object classifier in different attribute domains as follows,
%\begin{equation}
%G_{\mathcal{D}_{o}} = \mathbb{E}_{\mathcal{D}_o }\frac{\partial l(\varphi_a(\hat{z}_a^o;\theta_{a}),y_a )}{\partial \theta_{a} },
%\end{equation}
%\begin{equation}
%G_{\mathcal{D}_{\bar{o}}} = \mathbb{E}_{\mathcal{D}_{\bar{o}} }\frac{\partial l(\varphi_a(\hat{z}_a^{\bar{o}};\theta_{a}),y_a)}{\partial \theta_{a}}.
%\end{equation}

\begin{equation}
G_{j}^i = \frac{\partial l(\varphi_j(\hat{z}_j^{i};\theta_{j}),y_j^i)}{\partial \theta_{j}}.
\end{equation}

The gradient represents the optimal path. It is easier to obtain invariant predictions in different domains by encouraging the same optimization paths in all domains~\cite{shi2021gradient}.
In order to align different domains at the gradient level and learn the invariance associated with label, we penalize the domain prediction ability by minimizing the Euclidean distance $d_{euc}(\cdot, \cdot)$ between the two gradients as shown below,
\begin{equation}
% L_{grad^a}=d_{Eucl}(G_{\mathcal{D}_{o}},G_{\mathcal{D}_{\bar{o}}} ).
 L_{grad}=d_{euc}(G_{attr}^{(a,o)},G_{attr}^{(a,\bar{o})})+d_{euc}(G_{obj}^{(a,o)},G_{obj}^{(\bar{a},o)}).
 \end{equation}

We measure the alignment by calculating the Euclidean distances of gradients across different domains. 
In addition, cosine distance is also considered to measure the alignment of domains in the ablation experiment (see Section{\bf~\ref{ablation}}). 

%Consider the case that recognize object $o$ over the two attribute domains $\mathcal{D}_{attr}=\left \{ \mathcal{D}_{a}, \mathcal{D}_{a^*} \right \} $ with $x_{a,o}$ and $x_{\bar{a},o}$ as inputs.

%We discard channels in the representations that are specific to the attribute domain and compute the cross entropy loss for the remaining object-independent features.
%\begin{equation}
%L_{rep}^o=l(\varphi_o(\hat{z}_{\mathcal{D}_{attr}};\theta_o ),y_o).
%\end{equation}

%Similarly, the attribute-irrelevant invariance of object at the gradient level can be obtained by reducing the Euclidean distance of the gradient between the two attribute domains,
%\begin{equation}
% L_{grad}^o=d_{Eucl}(G_{\mathcal{D}_{a}},G_{\mathcal{D}_{\bar{a}}} ).
%\end{equation}

By introducing these regularizing terms, 
we can adaptively look for domain-specific channels and discard them, forcing the network to find an invariant relationship between the input image and the label at the representation-level consistency.
We also conduct all the domains optimized in the same direction at the gradient-level consistency.
Finally, we get decoupled attribute features and object features, which will improve the predictive performance of the model in unseen compositions.

\subsection{Training and Inference}
For training, we borrow from previous works using the composition classification loss in embedding learning to explore the dependence between attributes and objects,
\begin{equation}
L_{comp}=\sum_{i \in \mathcal{C}}h_{comp}(x^i,y^i).
\end{equation}

Simultaneously, we employ invariant feature learning to decouple attributes and objects to explore their independence.
Finally, the objective of optimization can be expressed as,
\begin{equation}
%L=L_{com}+L_{cls}^i+\lambda_1*L_{rep}^i+\lambda_2*L_{grad}^i,\ i=\left\{a,o\right\},\label{con:overall}
L=L_{comp} + L_{cls} + \lambda_1 L_{rep} + \lambda_2 L_{grad},\ \label{con:overall}
\end{equation}
where $\lambda_{1}$ and $\lambda_{2}$ are trade-off parameters.

During inference, given an image in the test set, we project it into the common embedding space. The distance between visual embedding features and all candidate pair vectors is calculated and sorted to obtain a pair score predicted in the form of coupling.
On the other hand, we use classifiers to predict attributes and objects separately in a decoupled manner and combine the predicted results into a pair score. The final prediction result is obtained by adding the two pair scores, which will improve the performance of the model for both seen and unseen pairs.

\section{Experiments}
\subsection{Datasets}
Mit-States~\cite{isola2015discovering} and UT-Zappos50K~\cite{yu2014fine} are two benchmark datasets widely used in CZSL task.

After careful observation of the dataset, we also discover three significant problems. 
First, because Mit-States is labelled automatically using early image search engine technology~\cite{atzmon2020causal}, it contains much noise. For example, there is an image labelled \emph{pierced bear}, but it is actually a brown ceramic pot. Second, the existence of both super-classes and sub-classes in this dataset, such as \emph{animal} and \emph{horse}, as well as \emph{fruit} and \emph{apple}, can create ambiguity.
Thirdly, the semantic expression of some attributes is not clear enough. For example, \emph{big bear} and \emph{large bear} are precisely the same from the picture. 
In light of these issues, we believe that the Mit-States dataset is too noisy to evaluate effectively. Therefore, we use UT-Zappos50K, Clothing16K, and AO-CLEVr for the experiment. % clothing数据集名称

UT-Zappos$50$K~\cite{yu2014fine} is a fine-grained shoes dataset which contains about $33$k images with $12$ object classes and $16$ attribute classes. The object concepts are mainly the types of shoes (e.g. heels, slippers), while the attribute concepts are mainly the material of shoes (e.g. canvas, leather).
Following the generalized evaluation protocol proposed by~\cite{2019Task}, we test on both seen and unseen pairs.
We adopt the standard split from~\cite{2019Task,2021Open}, the training set has about $23$k images belonging to $83$ attribute-object pairs. The validation set has about $3$k images consisting of $15$ seen pairs and $15$ unseen pairs. And the test set has about $3$k images consisting of $18$ seen pairs and $18$ unseen pairs.

Clothing16K\footnote{\url{https://www.kaggle.com/kaiska/apparel-dataset}} was initially a dataset used for multi-label classification in Kaggle competitions with $8$ object classes and $9$ attribute classes. The object concepts are mainly the types of clothing (e.g. shirt, pants), while the attribute concepts are mainly the clothing colour(e.g. black, green).
We find that the attributes and objects of this dataset are very distinct and almost contain no noise, which is very suitable for the CZSL task. Therefore, we split the dataset by ourselves following the generalized ZSL principle~\cite{2019Task}.
The training set has about $7$k images in $18$ attribute-object pairs. The validation set consists of $10$ seen pairs and $10$ unseen pairs with a total of about $5$k images. And the test set consists of $9$ seen pairs and $8$ unseen pairs with a total of about $3$k images.

AO-CLEVr~\cite{atzmon2020causal} is a synthetic dataset consisting of $3$ object classes (e.g. sphere, cube, cylinder) and $8$ attribute classes (e.g. yellow, gray), with $24$ compositional classes in total. 
We also split the dataset following the generalized ZSL principle~\cite{2019Task}.
The training set has about $103$k images in $16$ attribute-object pairs. The validation set consists of $4$ seen pairs and $4$ unseen pairs with a total of about $39$k images. And the test set has about $38$k images from $4$ seen pairs and $4$ unseen pairs.

\subsection{Metric}
Following~\cite{2019Task,2021Open}, we test the performance by the accuracy of their top-1 prediction for recognizing 
% attribute (\emph{Attr}), object (\emph{Obj}), 
seen pairs (\emph{Seen}) and unseen pairs (\emph{Unseen}) in the validation set and test set. 
To account for the inherent bias towards seen pairs, we follow Chao et al.~\cite{chao2016empirical} to add a calibration bias term to the unseen pairs to balance the seen-unseen accuracy. 
When the calibration value is positive, the prediction accuracy of the unseen pair will be high, and when the calibration value is negative, the model tends to have a bias towards seen pairs.
As the candidate value changes, a curve can be drawn with the accuracy of seen pairs on the X-axis and unseen pairs on the Y-axis.
We report the Area Under Curve (\emph{AUC}) to evaluate the overall performance. 
We also consider the best harmonic mean (\emph{HM}) of seen accuracy and unseen accuracy defined as $2(Seen*Unseen)/(Seen+Unseen)$ in this curve, which can penalize the large performance discrepancies between two quantities and as such enables the model to verify performance on both seen and unseen pairs simultaneously.

\subsection{Implementation Details}
Following~\cite{2019Task,2021Open}, we use ResNet-18~\cite{he2016deep} pretrained on ImageNet~\cite{deng2009imagenet} as the feature extractor. For a fair comparison with prior works, we do not finetune this network. 
% We use the $512$-dimension output of the last fully connected layer as the input visual feature.
The extracted $512$-dimension features are mapped into a common embedding space through an image embedding function consists of $2$ fully-connected layers.
Then, we build an attribute disentangler and an object disentangler with a fully-connected layer to map the features into attribute subspace and object subspace respectively. Finally, an attribute classifier and an object classifier implemented by a fully-connected layer are trained to recognize concepts respectively.
Simultaneously, we map concatenated compositional text features into the common embedded space.
% We initialize the text embedding functions with $300$-dimensional word2vec embeddings for all datasets. 
% After concatenating the text features of attribute and object, a projection function map the concatenated features into the common embedding space with a fully-connected layer.
% All experimental models are trained for $100$ epochs, using ADAM as an optimizer. The initial learning rate is set to $0.001$ and the weight decay is set to $5 \times 10^{-5}$. 
We use Adam optimizer with a initial learning rate set to $0.001$ and a weight decay set to $5 \times 10^{-5}$.
The $\lambda_{1}$ and $\lambda_{2}$ in Eqs.~\eqref{con:overall} are respectively set to $1$ and $10$ in all experiments.

\subsection{Compared Methods}
We compare our work with several methods. 

(1) LE+~\cite{2018Attributes} uses GloVe~\cite{pennington2014glove} word vectors to represent attribute and object concepts and trains the neural network to project the concatenated concept features and visual features to a joint embedding space.

(2) AttrAsOp~\cite{2018Attributes} treats the attribute as a matrix operator and treats the object as a vector. Then conducts attribute-conditioned transformations to learn unseen attribute-object pairs.

(3) SymNet~\cite{2020Symmetry} considers the symmetry principle in the attribute-object composition process and introduces group theory as a foundation for axiomatics.

(4) TMN~\cite{2019Task} trains a set of network modules jointly with a gating network where the compositional reasoning task is divided into sub-tasks that multiple small networks can solve in a semantic concept space. 

(5) CompCos~\cite{2021Open} proposes an open world setting where all the compositions of attributes and objects could potentially exist. A feasible strategy is proposed to remove the impossible compositions.

(6) VisProd~\cite{lu2016visual}. Unlike the above methods, VisProd does not model the composition explicitly but imposes attribute classifier and object classifier independently over the image features. The prediction result of a composition is the product of the probability of each element: $P(c)=P(a)\times P(o)$.

\subsection{Quantitative Result}
We summarize the results for our method and other methods on the three datasets in Table~\ref{tab:results12}.
Our method outperforms almost all reported results.
% Compared with the state-of-the-art, our method increases the classification accuracy by $5.6\%$ (Attr), $5.3\%$ (Unseen), $8.4\%$ (HM), and $8.0\%$ (AUC) on the UT-Zappos50K dataset. And on the Clothing16K dataset, our method increases the classification accuracy by $2.9\%$ (Obj), $1.7\%$ (Unseen), $3.1\%$ (HM), and $2.7\%$ (AUC). 
Compared with the accuracy of seen pairs, our method improves the accuracy of unseen pairs to a greater extent.
This is because our method inevitably loses the spurious correlation between attributes and objects while learning them independently. In other words, it hurts the model's bias against the seen pairs.
Although the ability of model to recognize seen pairs is weak, HM and AUC, the metrics of comprehensive recognition ability, increased. 
The experimental result sufficiently proves the superiority of our proposed method.

\begin{table*}[ht]
\begin{center}
\setlength\tabcolsep{0.8pt}         % 列间距
\caption{Comparative experiment between recent methods with our method on UT-Zappos50K, Clothing16K, and AO-CLEVr.} \label{tab:results12}
\begin{tabular}{lcccc|cccc|cccc}
% \hline
  &  \multicolumn{4}{c}{UT-Zappos50K} &  \multicolumn{4}{c}{Clothing16K} &
  \multicolumn{4}{c}{AO-CLEVr} \\
  \cline{2-13}\noalign{\smallskip}
  Method & Seen & Unseen & HM & AUC &  Seen & Unseen & HM & AUC & Seen & Unseen & HM & AUC\\
\hline
LE+~\cite{2018Attributes}  & $53.0$ & $61.9$  & $41.0$ & $25.7$  & $93.9$ & $88.3$  & $77.4$ & $76.0$ & $95.7$ & $99.2$  & $92.3$ & $93.5$\\
AttrAsOp~\cite{2018Attributes} & $59.8$ & $54.2$  & $40.8$ & $25.9$ & $95.1$ & $80.1$  & $60.0$ & $58.7$ & $95.5$ & $85.5$ & $64.8$ & $65.8$\\
SymNet~\cite{2020Symmetry}  & $49.8$ & $57.4$  & $40.4$& $23.4$& $95.7$ & $90.2$  & $73.4$& $75.2$  & $87.1$ & $97.8$ & $71.8$ & $74.2$\\
VisProd~\cite{lu2016visual} & $56.6$ & $60.2$ & $43.7$ & $28.1$ & $96.4$ & $91.4$  & $74.7$ & $77.5$ & $91.9$ & $98.2$ & $71.3$ & $75.6$\\
TMN~\cite{2019Task}  & $58.7$ & $60.0$  & $45.0$& $29.3$ & $94.9$ & $89.7$  & $80.9$& $79.5$ & $96.1$ & $93.9$  & $86.9$& $87.1$ \\
CompCos~\cite{2021Open}  & ${\bf59.8}$ & $62.5$  & $43.1$ & $28.7$& $96.9$ & $93.0$  & $83.9$ & $84.7$ & $96.3$ & $99.1$  & $94.5$ & $94.2$\\
\hline
Ours  & $56.9$ & ${\bf65.5}$ & ${\bf46.2}$ &${\bf30.6}$& ${\bf96.9}$ & ${\bf94.6}$ & ${\bf86.3}$ &${\bf87.0}$ & ${\bf97.1}$ & ${\bf 99.3}$ & ${\bf 95.1}$ & ${\bf 95.6}$ \\
\hline
\end{tabular}
\end{center}
\end{table*}

\begin{figure}[t]
\centering
\includegraphics[height=4.5cm]{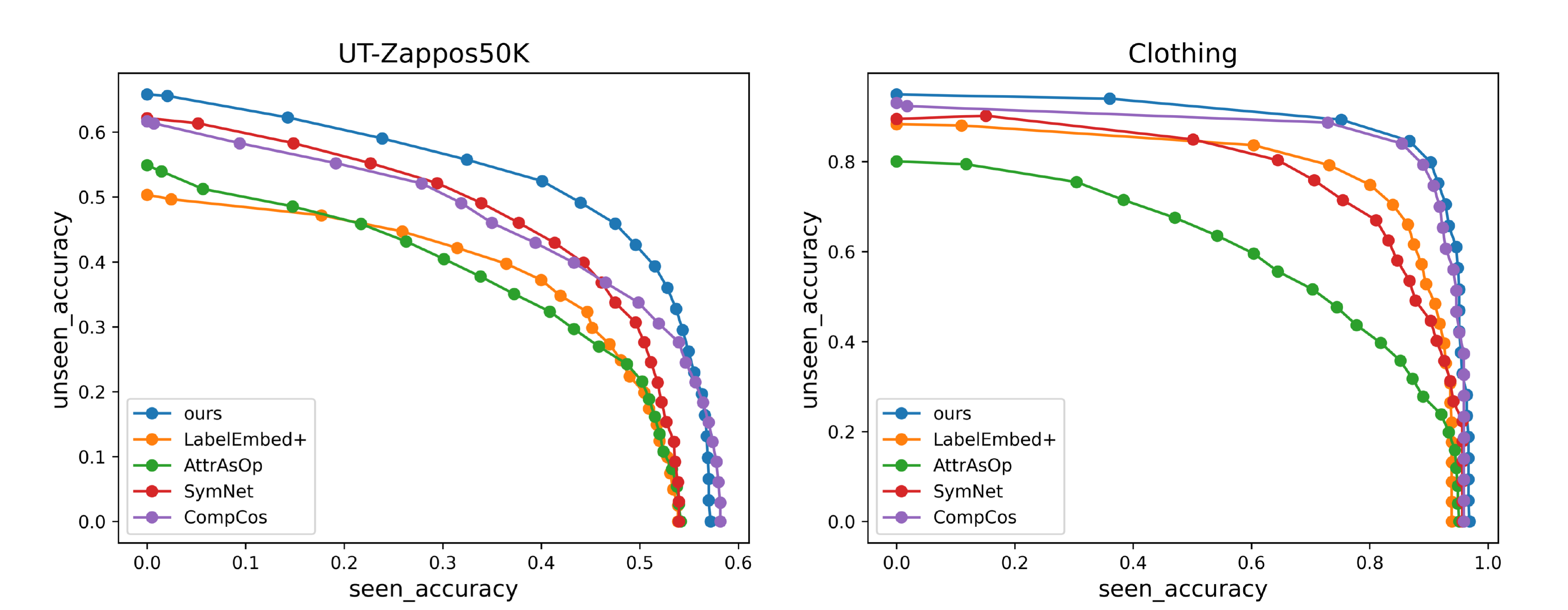}   % 4.7
\caption{Unseen-seen accuracy on UT-Zappos50K and Clothing16K under various calibration biases.
}
\label{fig:curve}
\end{figure}

% \vspace{-10mm}
Figure~\ref{fig:curve} shows the unseen-seen accuracy curve on the UT-Zappos50K and Clothing16K dataset. With the increase of calibration value, the classification accuracy of seen pairs decreases while that of unseen pairs increases. 
This is a general and essential trade-off when learning models that are robust for interventions~\cite{rothenhausler2021anchor}.
Compared to other methods, our method keeps a better balance between seen and unseen pairs on both datasets, which leads to better performance.

Overall, the results on these challenging datasets strongly support our idea of leveraging invariant mechanisms to decouple attributes and objects effectively. 
Learning attributes and objects in a decoupled way may discourage certain types of correlations~\cite{atzmon2020causal}, so the model can not benefit from them when the test and training distributions are the same, that is, recognizing seen pairs. However, when recognizing unseen pairs, where the test and training distributions are different, our method of improving generalization performance can come into play without taking advantage of these spurious correlations.

% \vspace{-6mm}
\begin{table*}[ht]
\begin{center}
\caption{Analysis of each component on UT-Zappos50K and Clothing16K.} \label{tab:results4}
\begin{tabular}{l c c c c | c c c c}
& \multicolumn{4}{c}{UT-Zappos50K} & \multicolumn{4}{c}{Clothing16K}\\
\cline{2-9}\noalign{\smallskip}
  \multicolumn{1}{c}{Method} & Seen & Unseen & HM & AUC & Seen & Unseen & HM & AUC\\
\hline\noalign{\smallskip}
$L_{comp}$ & ${\bf58.4}$ & $58.0$ & $43.5$ & $27.8$ & $96.9$ & $91.8$ & $81.6$ & $82.9$\\
$L_{cls}$ & $56.0$ & $63.5$ & $44.0$ & $27.7$ & $95.1$ & $93.5$ & $82.2$ & $83.2$\\
$L_{cls}$+$L_{comp}$ & $57.0$ & $63.4$ & $44.2$ & $28.8$ & $96.2$ & $93.7$ & $84.7$ & $84.8$ \\
$L_{cls}$+$L_{comp}$+$L_{rep}$ & $55.9$ & $65.5$ & $45.3$ & $29.8$ & $96.7$ & $94.0$ & $85.3$ & $85.3$ \\
$L_{cls}$+$L_{comp}$+$L_{grad}$ & $56.6$ & $64.4$ & $46.1$ & $30.0$ & ${\bf97.2}$ & $94.2$ & $85.6$ & $86.3$\\
$L_{cls}$+$L_{comp}$+$L_{rep}$+$L_{grad}$  & $56.9$ & ${\bf65.5}$ & ${\bf46.2}$ &${\bf30.6}$ & $96.9$ & ${\bf94.6}$ & ${\bf86.3}$ & ${\bf87.0}$ \\
\hline
\end{tabular}
\end{center}
\end{table*}

% \vspace{-10mm}
\begin{table*}[ht]
\begin{center}
\caption{Analysis of parameter $\alpha$ on UT-Zappos50K and Clothing16K.} \label{tab:results5}
\begin{tabular}{c c c c c | c c c c}
& \multicolumn{4}{c}{UT-Zappos50K} & \multicolumn{4}{c}{Clothing16K}\\
\cline{2-9}\noalign{\smallskip}
   $\alpha$  & Seen & Unseen & HM & AUC & Seen & Unseen & HM & AUC\\
\hline\noalign{\smallskip}
1/6 & $56.9$ & ${\bf65.5}$ & ${\bf46.2}$ &${\bf30.6}$ & $96.8$ & $94.4$ & $86.4$ & $86.6$\\
1/4 & $56.3$ & $65.0$  & $45.5$ & $29.9$ & ${\bf96.9}$ & ${\bf94.6}$ & $86.3$ & ${\bf87.0}$\\
1/3 & ${\bf57.5}$ & $63.4$  & $45.2$ & $29.2$ & $96.8$ & $94.5$ & $86.4$ & $86.6$ \\
1/2 & $53.6$ & $65.4$  & $44.4$ & $28.5$ & $96.8$ & $93.9$ & ${\bf86.7}$ & $86.5$ \\
\hline
\end{tabular}
\end{center}
\end{table*}

% \vspace{-10mm}
\begin{table*}[ht]
\begin{center}
\caption{Analysis of distance function on UT-Zappos50K and Clothing16K.} \label{tab:results6}
\begin{tabular}{c c c c c | c c c c}
& \multicolumn{4}{c}{UT-Zappos50K} & \multicolumn{4}{c}{Clothing16K}\\
\cline{2-9}\noalign{\smallskip}
Distance Function & Seen & Unseen & HM & AUC& Seen & Unseen & HM & AUC\\
\hline\noalign{\smallskip}
Euclidean & $56.9$ & ${\bf65.5}$ & ${\bf46.2}$ &${\bf30.6}$ & ${\bf96.9}$ & ${\bf94.6}$ & ${\bf86.3}$ & ${\bf87.0}$\\
Cosine & ${\bf58.3}$ & $62.6$  & $44.2$ & $28.6$ & $96.7$ & $94.3$  & $85.0$ & $85.9$\\
\hline\noalign{\smallskip}
\end{tabular}
\end{center}
\end{table*}

% \vspace{-10mm}
\subsection{Ablation Study}~\label{ablation}
To verify the effect of each proposed component, we conduct ablation experiments on the UT-Zappos50K and Clothing16K datasets.
As shown in Table~\ref{tab:results4}, when only compositional classification loss (denoted as “$L_{comp}$”) is applied, the model have a positive bias towards the seen pairs because of the dependence between objects and attributes. 
When the concepts are learned in a decoupled way using attribute and object classifiers (denoted as “$L_{cls}$”), the model is biased towards unseen pairs since the correlation between attributes and objects is removed.
The utilization of representation invariant mechanism (denoted as “$L_{rep}$”) can help the model to discard domain-specific spurious features at the representation level, thus improving the performance of the model.
When the gradient invariant mechanism (denoted as “$L_{grad}$”) is employed, the gradients of different domains are optimized in the same direction. 
Through these two invariant learning mechanisms, the model can learn the optimal attribute classifier and object classifier, which remarkably improves the comprehensive performance of the model.

{\bf Effect of parameter $\alpha$}. The scale parameter $\alpha$ is employed to control the proportion of discarding in Eqs.~\eqref{con:mask}.  
We select $\alpha$ in $\left \{ \frac{1}{6},\frac{1}{4} ,\frac{1}{3} ,\frac{1}{2}   \right \} $ and report the performance of the model in Table~\ref{tab:results5}.
For the UT-Zappos50K dataset, the optimal performances can be observed when $\alpha$ is set to $\frac{1}{6}$.
For the Clothing16K dataset, the optimal performances can be observed when $\alpha$ is set to $\frac{1}{4}$.
A suitable $\alpha$ can subtly discard domain-specific features and help the model generalize from known concepts to unseen ones by using domain-invariant features.

{\bf Effect of distance function}. In the gradient invariant mechanism, we use Euclidean distance to measure the distance between gradients in different domains.
In addition, our method also works with cosine distance. As shown in Table~\ref{tab:results6}, the performance of Euclidean distance is better than cosine distance, probably because we pay more attention to the absolute numerical differences between gradients.

\begin{figure}[t]
\centering
\includegraphics[height=4.2cm]{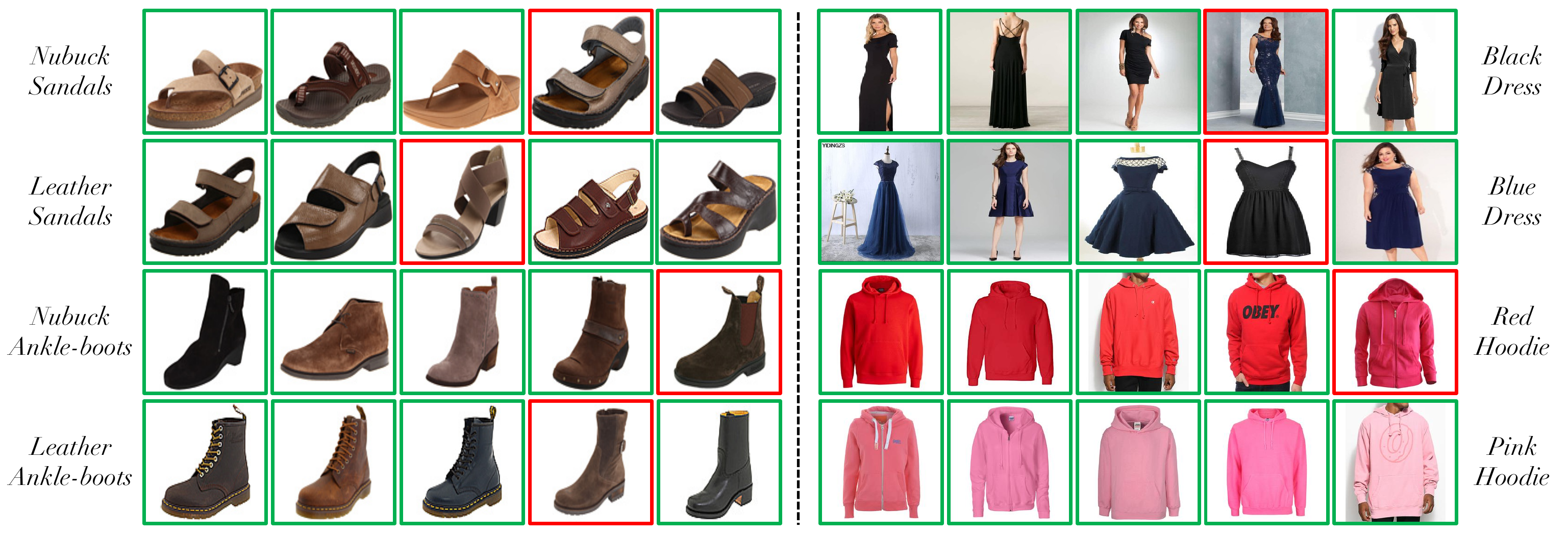}
\caption{Qualitative results of retrieving \emph{nubuck sandals}, \emph{leather sandals}, \emph{nubuck ankle-boots}, \emph{leather ankle-boots} in UT-Zappos50K and \emph{black dress}, \emph{blue dress}, \emph{red hoodie}, \emph{pink hoodie} in Clothing16K.  
}
\label{fig:visual}
\end{figure}

\subsection{Image Retrieval}
To qualitatively evaluate our method, we further report image retrieval results.
Figure~\ref{fig:visual} shows examples of retrieving images. 
The query is made up of attribute text and object text. 
We choose compositions of different objects with the same attribute and compositions of different attributes with the same object.
For UT-Zappos50K and Clothing16K datasets, our method can retrieve a certain number of correct samples in the top-$5$, indicating that our method can solve the combinatorial generalization problem.

\section{Conclusions}
In reality, there are many situations where data distribution is different during training and testing.
Inspired by the idea of exploring domain invariance in the DG task, we propose the representation invariant mechanism and gradient invariant mechanism to find essential features of attributes and objects, and finally learn attribute and object classifiers that can be generalized to any new composition.
The limitation of our method is that it can be challenging to decouple attributes or objects when they can only form one composition in the training set. At this point, the model is more likely to overfit to the seen pairs. 
In the future, we will delve into studying the core features of such concepts. Besides, we will also explore the application of generalization ideas to multiple sub-concept composition scenarios and even other avenues of research.

\section*{Acknowledgement}

This work was supported in part by National Natural Science Foundation of
China (NSFC) No. 61922015, 62106022, U19B2036, and in part by Beijing Natural Science
Foundation Project No. Z200002. 

\clearpage
\bibliographystyle{splncs04}
\bibliography{egbib}
\end{document}